\documentclass[letterpaper, 10 pt, journal, twoside]{ieeeconf}
\IEEEoverridecommandlockouts                              %
\usepackage{graphicx,amsmath,amssymb,bm,xcolor,soul,nicefrac,listings,dsfont,caption,subcaption,wrapfig,sidecap} 
\usepackage[hidelinks]{hyperref}
\usepackage{courier}
\usepackage[algo2e]{algorithm2e} 

\SetCommentSty{mycommfont}
\usepackage{algorithmic}
\usepackage{booktabs}
\usepackage{siunitx}
\usepackage{algorithm}

\newcommand{\trsp}{{\scriptscriptstyle\top}}

\title{\LARGE \bf
From Movement Primitives to Distance Fields to Dynamical Systems
}

\author{Yiming Li and Sylvain Calinon
\thanks{The authors are with the Idiap Research Institute, Martigny, Switzerland and also with EPFL, Lausanne, Switzerland. (name.surname@idiap.ch)}}%

\begin{document}

\maketitle
\thispagestyle{empty}
\pagestyle{empty}

\begin{abstract}
Developing autonomous robots capable of learning and reproducing complex motions from demonstrations remains a fundamental challenge in robotics. On the one hand, movement primitives (MPs) provide a compact and modular representation of continuous trajectories. On the other hand, autonomous systems provide control policies that are time independent. We propose in this paper a simple and flexible approach that gathers the advantages of both representations by transforming MPs into autonomous systems. 
The key idea is to transform the explicit representation of a trajectory as an implicit shape encoded as a distance field. This conversion from a time-dependent motion to a spatial representation enables the definition of an autonomous dynamical system with modular reactions to perturbation.
Asymptotic stability guarantees are provided by using Bernstein basis functions in the MPs, representing trajectories as concatenated quadratic Bézier curves, which provide an analytical method for computing distance fields. This approach bridges conventional MPs with distance fields, ensuring smooth and precise motion encoding, while maintaining a continuous spatial representation. By simply leveraging the analytic gradients of the curve and its distance field, a stable dynamical system can be computed to reproduce the demonstrated trajectories while handling perturbations, without requiring a model of the dynamical system to be estimated.
Numerical simulations and real-world robotic experiments validate our method's ability to encode complex motion patterns while ensuring trajectory stability, together with the flexibility of designing the desired reaction to perturbations.

An interactive project page demonstrating our approach is available at \url{https://idiap.github.io/mp-df-ds/}.
\end{abstract}


\section{Introduction and Related Work}

Learning complex motion skills by optimization or through demonstrations remains a fundamental challenge in robotics, which can computationally be addressed from different perspectives. Movement primitives (MPs) offer a compact and modular framework for encoding, generalizing, and reproducing demonstrated trajectories by superposition of basis functions, see \cite{Calinon19MM} for a review. Various MP formulations have been proposed, including dynamical movement primitives (DMPs) \cite{ijspeert2013dynamical,saveriano2023dynamic}, Gaussian mixture regression (GMR) \cite{calinon2007learning}, probabilistic movement primitives (ProMPs) \cite{paraschos_probabilistic_2013}, kernelized movement primitives (KMPs) \cite{huang2019kernelized}, or Fourier movement primitives (FMPs) \cite{Kulak20RSS}.

Conventional MPs are often limited in their ability to robustly handle perturbations and dynamically adapt to environmental changes, due to their explicit representation of the signals based on a time or phase variable. This downside has motivated the development of alternative autonomous systems formulations to provide time-independent dynamical system representations,  inherently robust to perturbations, at the expense of complexifying the problem of modeling long and/or high-dimensional movements. Examples include SEDS \cite{khansari2011learning}, CLF-DM \cite{khansari2014learning}, Imitation Flow \cite{urain2020imitationflow}, and Neural Contractive Dynamical Systems (NCDS) \cite{beik2024neural}, which all ensure stability and adaptability by learning or imposing constraints on the system dynamics. 

\begin{figure}[t]
    \centering
    \includegraphics[width=\linewidth]{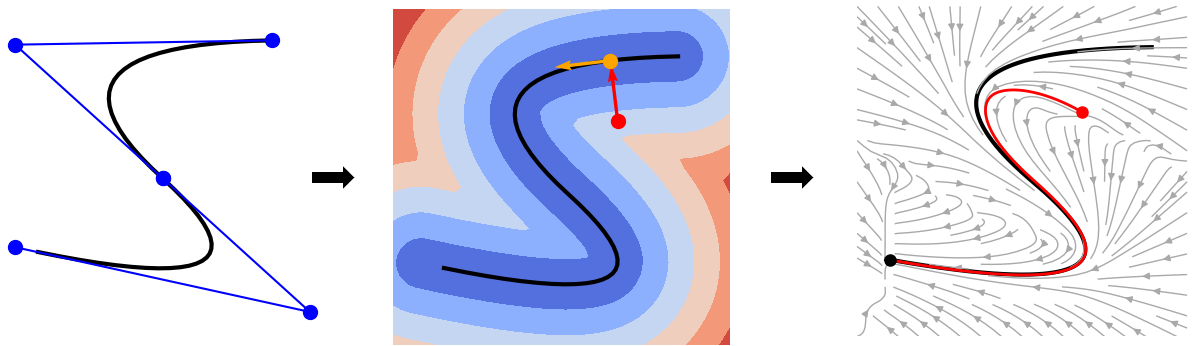}
    \caption{From movement primitive (\emph{left}) to distance field (\emph{middle}) to dynamical system (\emph{right}). \emph{Left:} A S-shape trajectory is parameterized using the concatenation of 2 quadratic curves. Control points are shown in blue. \emph{Middle:} Distance field of the quadratic spline. The gradient of the distance field at a given point (in red) is perpendicular to the curve at its closest point (in orange). This property is used to compute the distance field analytically and build the stable dynamical system (\emph{Right}). Red and black bullets are a random start point and the equilibrium point at the end of the trajectory.}
    \label{fig:combined}
    \vspace{-6mm}
\end{figure}

We propose in this paper a simple and flexible approach that gathers the advantages of both representations by transforming MPs into autonomous systems. 
Our work follows the same foundational principle of using autonomous systems for robust motion generation, but differs by avoiding learning-based modeling altogether. Instead, we directly leverage the geometric properties of a reference trajectory encoded as a distance field, allowing us to define an autonomous system with asymptotic stability guarantees through analytic computation. A related concept has been explored in contact-aware tasks such as robotic polishing, where surfaces and normal vectors guide the design of dynamical systems \cite{amanhoud2019dynamical}. We apply the idea to full trajectory encoding beyond surface interaction. 


The core idea is to reinterpret the demonstrated trajectories as implicit spatial representations encoded as distance fields. Rather than representing motion as a sequence of time-parameterized states, we encode the shape of the trajectory in space. The resulting field defines the shortest distance to the reference trajectory at each spatial location, enabling motion generation as gradient descent over this field.

Asymptotic stability is ensured by incorporating Bernstein basis functions into movement primitives (MPs), representing trajectories as concatenated quadratic Bézier curves. This formulation enables the analytical computation of distance fields, eliminating the need for point sampling or heuristic weighting \cite{chen2021closed}. By bridging conventional MPs with distance fields, this approach provides smooth and precise motion encoding within a continuous spatial representation. Leveraging the analytical gradients of both the curve and its distance field, a stable dynamical system can be directly derived to reproduce demonstrated trajectories and handle perturbations, without requiring explicit system modeling or learning techniques.

The motivation of representing structure implicitly in continuous space is inspired by recent advances in Signed Distance Fields (SDFs) and Neural Radiance Fields (NeRFs) \cite{park2019deepsdf,mildenhall2021nerf}, offering compact and continuous representations for shapes and scenes. These techniques have influenced robotics, particularly in motion planning and scene understanding \cite{ratliff2009chomp,li2024representing}. Our work in this paper instead focuses on representing 1D trajectories embedded in high-dimensional spaces. By exploiting this lower-dimensional structure, we derive closed-form distance fields from concatenated quadratic Bézier curves, enabling smooth, precise, and efficient computation of spatial gradients for motion generation.

This spatial formulation offers several advantages. It eliminates the need for time synchronization, naturally supports modular responses to perturbations, and provides a continuous and robust encoding of motion by avoiding the discretization artifacts common in numerical methods.

The main contributions of this paper are:
\begin{itemize}
    \item We propose a novel use of distance fields as an implicit trajectory representation for autonomous systems, shifting from conventional time-dependent trajectory encoding to a spatial formulation guided by distance gradients.
    \item We derive an analytical method for computing distance fields to reference trajectories represented by concatenated quadratic Bézier curves. This approach enables efficient and accurate computation of distance and gradient fields while preserving the ability to represent complex motion patterns.
    \item We show that combining movement primitives with their corresponding gradient fields naturally yields autonomous dynamical systems with provable asymptotic stability—without requiring system modeling or learning-based estimation.
\end{itemize}

We validate our approach through numerical simulations and real-world robot experiments, demonstrating its effectiveness in encoding complex motions, generating stable and smooth trajectories, and adapting to external perturbations. Additionally, we highlight the simplicity and flexibility of our framework in:
\begin{itemize}
\item integrating multiple demonstrations through simple field operations;
\item scaling to high-dimensional settings such as robot joint spaces;
\item enabling adaptive motions that achieve a balance between robustness to perturbations and generalization to new trajectories.
\end{itemize}

\section{Representing Trajectories as Distance Fields}


Given a reference trajectory \(\mathbf{F}\) whose points are given by \(\mathbf{f}(t): [0, T] \to \mathbb{R}^D\) parameterized by \(t \in [0, T]\), where \(T\) is the trajectory length and \(D\) is the spatial dimension, the distance field \(d(\mathbf{x})\) at any point \(\mathbf{x} \in \mathbb{R}^{\textcolor{black}{D}}\) is defined as:
\begin{equation}
    d(\mathbf{x}) = \min_{t \in [0, T]} \|\mathbf{f}(t)-\mathbf{x}\|,
\end{equation}
where \(\|\cdot\|\) denotes the Euclidean norm. This function assigns each point in space a scalar value corresponding to its shortest distance to the trajectory. The gradient of the distance field, \(\mathbf{\nabla d}(\mathbf{x})\), provides the direction of steepest ascent. Given a point \(\mathbf{x}\), the closest point on the trajectory is given by
\begin{equation}
    \label{eq:proj}
    \mathbf{x}_{\text{proj}} = \mathbf{x} - d(\mathbf{x}) \mathbf{\nabla d}(\mathbf{x}).
\end{equation}
In other words, one obtains the projection by subtracting from \(\mathbf{x}\) the displacement \(d(\mathbf{x}) \mathbf{\nabla d}(\mathbf{x})\), which points from the trajectory toward \(\mathbf{x}\). The gradient \(\mathbf{\nabla d}(\mathbf{x})\) has the unit norm \(\|\mathbf{\nabla d}(\mathbf{x})\| = 1\).

While distance fields can be evaluated numerically by discretizing the trajectory and evaluating the minimum distance from sampled points, this approach can introduce discretization errors and result in non-smooth transitions between points. Instead, we present an analytical distance computation by representing the trajectory as Bernstein basis functions.

\subsection{Quadratic spline as Movement Primitives}  
Splines are widely used as movement primitives due to their ability to provide smooth and flexible trajectory representations. We use here the term \emph{quadratic spline} to define the concatenation of quadratic Bézier curves, corresponding to piecewise Bernstein polynomial functions that ensures continuity in both position and velocity. \textcolor{black}{A quadratic spline is composed of \(N\) segments. Each segment \(\mathbf{f}_i\) is a quadratic Bézier curve parameterized by a local time \(t_i \in [0,1]\):}
\begin{equation}
\mathbf{f}_i(\textcolor{black}{t_i}) = (1-\textcolor{black}{t_i})^2, \mathbf{w}^{1}_i + 2(1-\textcolor{black}{t_i})\textcolor{black}{t_i}, \mathbf{w}^{2}_i + \textcolor{black}{t_i}^2, \mathbf{w}^{3}_i, \quad \textcolor{black}{t_i} \in [0,1],
\end{equation}
where \(\mathbf{w}^{1}_i\), \(\mathbf{w}^{2}_i\), and \(\mathbf{w}^{3}_i\) are the control points of the \(i\)-th segment. \textcolor{black}{\textcolor{black}{The full trajectory \(\mathbf{f}(t)\) is formed by mapping a global time \(t \in [0,T]\) to the correct segment \(i\) and its local time parameter \(t_i\). This is achieved by assuming each of the \(N\) segments has an equal duration, allowing the global time \(t\) to be linearly scaled to identify the active segment and its corresponding local time in the [0,1] interval.}} Quadratic splines can approximate complex trajectories by decomposing the motion into localized polynomial segments. To ensure smooth concatenation between consecutive segments, continuity constraints are imposed with
\begin{equation}
    \mathbf{w}^{3}_i = \mathbf{w}^{1}_{i+1},\quad
    \mathbf{w}^{2}_i - \mathbf{w}^{3}_i = \mathbf{w}^{1}_{i+1} - \mathbf{w}^{2}_{i+1},\quad
    \mathbf{w}^2_N = \mathbf{w}^3_N.
\end{equation}
The first constraint ensures positional continuity at the segment boundaries. The second guarantees consistent gradients, maintaining \(C^1\) continuity across the trajectory. The last enforces zero velocity at the end of the trajectory by setting the last two control points of the final Bézier segment to be equal so that the derivative of the curve—and hence the velocity—vanishes at the endpoint.

This formulation can be compactly expressed in matrix form as
\begin{equation}
\label{eq:qs}
    \mathbf{f}(t) = \bm{\phi}(t)\,\mathbf{w},\quad \bm{\phi}(t) = \mathbf{T}(t)\,\mathbf{B}\,\mathbf{C},
\end{equation}
where \(\mathbf{f}(t)\) represents the concatenated quadratic splines, \(\mathbf{T}(t) = [1, t, t^2]\) encodes the quadratic polynomial basis, \(\mathbf{B}\) is the corresponding coefficient matrix, and \(\mathbf{C}\) enforces the continuity constraints. \textcolor{black}{The basis function matrix \(\bm{\phi}(t)\) is a piecewise function of the global time \(t\) that internally handles the selection and time-scaling for the appropriate segment.} A detailed derivation of this formulation can be found in \cite{RCFS}.

The parameters \(\mathbf{w}\) determine the shape of the quadratic spline and are estimated to best fit a given reference trajectory. Given a set of trajectory points \(\{\mathbf{p}_k\}_{k=1}^{K}\) sampled at time instances \(\{t_k\}_{k=1}^{K}\), we formulate the parameter estimation as an optimization problem that minimizes the discrepancy between the spline representation and the reference trajectory \(\bm{f}(t)\). Specifically, we solve
\begin{equation}
    \min_{\mathbf{w}} \sum_{k=1}^{K} \left\| \mathbf{f}(t_k) - \mathbf{p}_k \right\|^2,
\end{equation}
where \(\mathbf{f}(t)\) is the spline-based reconstruction of the trajectory at \(t\), and \(\|\cdot\|\) denotes the Euclidean norm. The optimization problem can be solved efficiently using least squares. By expressing the trajectory in the matrix form
\begin{equation}
\mathbf{P} = \bm{\Phi}\,\mathbf{w},
\end{equation}
where \(\mathbf{P} = [\mathbf{p}_1, \dots, \mathbf{p}_M]^\top\) is the matrix of trajectory points and \(\bm{\Phi}\) is the corresponding basis function matrix evaluated at \(\{t_j\}\), the optimal parameters are obtained by solving 
\begin{equation}
	\mathbf{w} = (\bm{\Phi}^\top \bm{\Phi})^{-1}\bm{\Phi}^\top\, \mathbf{P},
    \label{eq:LSQ}
\end{equation}
where \((\bm{\Phi}^\top \bm{\Phi})^{-1}\bm{\Phi}^\top\) is the Moore-Penrose pseudoinverse of \(\bm{\Phi}\). 
The quadratic spline is then reconstructed using (\ref{eq:qs}) for a smooth, continuous-time trajectory representation.

\subsection{Distance Fields of Quadratic Splines}\label{sec:dfqs}


By minimizing $c(t_i)=\frac{1}{2}(\mathbf{f}_i(t_i)-\mathbf{x})^\trsp(\mathbf{f}_i(t_i)-\mathbf{x})$ w.r.t. $t_i$, we get the closest point $\mathbf{f}_i(t_i)$ to $\mathbf{x}$ on the curve. By differentiating $c(t_i)$ and equating to zero, we get
\begin{equation}
    (\mathbf{f}_i(t_i) - \mathbf{x})^\top \; \mathbf{\dot{f}}_i(t_i) = 0,
    \label{eq:orth}
\end{equation}
where \(\mathbf{\dot{f}}_i(t_i)\) denotes the derivative of the spline with respect to \(t_i\). We can observe in the above equation that the closest point on the curve satisfies the orthogonality condition between the residual vector $\mathbf{f}_i(t_i)-\mathbf{x}$ and the tangent vector of the spline $\mathbf{\dot{f}}_i(t_i)$.
Substituting the quadratic spline formulation
\begin{equation}
    \begin{aligned}
        \mathbf{f}_i(t_i) &= (1-t_i)^2\, \mathbf{w}^{1}_i + 2(1-t_i)t_i\, \mathbf{w}^{2}_i + t_i^2\, \mathbf{w}^{3}_i \\
        \mathbf{\dot{f}}_i(t_i) &= -2(1-t_i)\, \mathbf{w}^{1}_i + (2-t_i)\, \mathbf{w}^{2}_i + t_i\, \mathbf{w}^{3}_i,
    \end{aligned}
\end{equation}
in \eqref{eq:orth} results in a cubic equation
\begin{equation}
    \bm{\alpha^3}_{i}\, t_i^3 + \bm{\alpha^2}_{i}\, t_i^2 + \bm{\alpha^1}_{i}\, t_i + \bm{\alpha^0}_{i} = 0,
\end{equation}
where the coefficients \(\bm{\alpha^3}_{i}, \bm{\alpha^2}_{i}, \bm{\alpha^1}_{i}, \bm{\alpha^0}_{i}\) are determined by \(\mathbf{w}^{1}_i\), \(\mathbf{w}^{2}_i\), \(\mathbf{w}^{3}_i\), and \(\mathbf{x}\). Therefore, the roots of this cubic equation within the interval \([0,1]\) correspond to the candidate closest points on the spline. The minimal distance is then computed as
\begin{equation}
    d_i(\mathbf{x}) = \min \big\{ \|\mathbf{f}_i(t_i^*) - \mathbf{x}\| : t_i^* \in [0,1] \big\}.
\end{equation}
If no roots satisfy \(t_i^* \in [0,1]\), the closest point occurs at one of the segment boundaries, \(t_i=0\) or \(t_i=1\). The overall minimal distance to a trajectory composed of multiple segments is given by
\begin{equation}
    d(\mathbf{x})  = \min \big\{ d_i(\mathbf{x})  : i = 1, \dots, \textcolor{black}{N}\big\},
\end{equation}
where \(d_i\) is the minimal distance for the \(i\)-th segment and N is the toal number of segments.

\subsection{Fusing Multiple Trajectories}  

An advantage of representing motion using distance fields is the simple formulation to handle multiple demonstrations, achieved through the union operation of distance fields. Specifically, given a set of reference trajectories \(\{\mathbf{f}^m(t)\}_{m=1}^{M}\), each represented as an individual distance field \(d^m(\mathbf{x})\), the combined distance field is computed using the union operation
\begin{equation}
    d(\mathbf{x}) = \min_{m \in \{1, \dots, M\}} d^m(\mathbf{x}),
\end{equation}
ensuring that at each point \(\mathbf{x}\), the distance field retains the minimal distance to any of the given trajectories. 



\subsection{From Distance Fields to Dynamical Systems}  

In this section, we demonstrate how the distance field representation can be utilized to design stable dynamical systems in a structured and efficient manner. \textcolor{black}{ Given a reference trajectory encoded by quadratic splines, we use the approach described in Section \ref{sec:dfqs} to calculate the distance field \( d(\mathbf{x}) \), the closest point \(\mathbf{x}_{\text{proj}}\) on the trajectory, 
and the corresponding phase variable  \(t_{\text{phase}} =\mathbf{f}^{-1}(\mathbf{x}_{\text{proj}})\). We define the dynamical system}
\begin{equation}
\label{eq:ds}
    \dot{\mathbf{x}} = -\alpha\; \mathbf{\nabla d}(\mathbf{x}) + \beta\; \textcolor{black}{\dot{\mathbf{f}}(t_{\text{phase}})},
\end{equation}  
where \( \mathbf{\nabla d}(\mathbf{x}) \) is the gradient of the distance field, pointing away from the trajectory. \textcolor{black}{The term \(\dot{\mathbf{f}}(t_{\text{phase}})\) is the velocity vector of the reference trajectory evaluated at \(t_{\text{phase}}\), obtained from the analytic derivatives of the quadratic spline.} \( \alpha > 0 \), \( \beta > 0 \) are gain parameters that regulate attraction to the trajectory and movement along it, respectively.
The first term in the system ensures attraction toward the reference trajectory (if a perturbation occurs), while the second term produces smooth movement along the reference trajectory. To analyze the stability of this system, we introduce the following Lyapunov function
\begin{equation}
    V(\mathbf{x}) = \frac{1}{2} d^2(\mathbf{x}).
\end{equation}  
Taking the time derivative and using the property \( \dot{d}(\mathbf{x}) = \mathbf{\nabla d}(\mathbf{x})^\top \dot{\mathbf{x}} \), we obtain:  
\begin{equation}
\begin{aligned}
    \dot{V}(\mathbf{x}) &= d(\mathbf{x}) \dot{d}(\mathbf{x}) \\
     &= d(\mathbf{x}) \mathbf{\nabla d}(\mathbf{x})^\top (-\alpha \ \mathbf{\nabla d}(\mathbf{x}) + \beta \ \textcolor{black}{\dot{\mathbf{f}}(t_{\text{phase}})}) \\
     &= -\alpha \ d(\mathbf{x}) \|\mathbf{\nabla d}(\mathbf{x})\|^2 + \beta \ d(\mathbf{x}) \mathbf{\nabla d}(\mathbf{x})^\top \textcolor{black}{\dot{\mathbf{f}}(t_{\text{phase}})} \\
     &= -\alpha \ d(\mathbf{x}) + \beta \ d(\mathbf{x}) \mathbf{\nabla d}(\mathbf{x})^\top \textcolor{black}{\dot{\mathbf{f}}(t_{\text{phase}})}.
 \end{aligned}
\end{equation}  
Since \( d(\mathbf{x}) \geq 0 \), the first term in \( \dot{V}(\mathbf{x}) \) is always non-positive. To ensure \( \dot{V}(\mathbf{x}) \leq 0 \), we analyze the second term \( d(\mathbf{x}) \mathbf{\nabla d}(\mathbf{x})^\top \textcolor{black}{\dot{\mathbf{f}}(t_{\text{phase}})} \) in two cases:

\textcolor{black}{
\textbf{Case 1: Interior Point ($0 < t_{\text{phase}} < T$).} This means the projection point \(\mathbf{x}_{\text{proj}}\) lies in the interior of the trajectory, the orthogonality condition from \eqref{eq:orth} holds. This ensures that the gradient vector is perpendicular to the trajectory velocity vector, making their dot product zero:
\[ \mathbf{\nabla d}(\mathbf{x})^\top \dot{\mathbf{f}}(t_{\text{phase}}) = 0. \]}

\textcolor{black}{
\textbf{Case 2: Boundary Points ($t_{\text{phase}} = 0$ or $t_{\text{phase}} = T$).}
\begin{itemize}
    \item At the endpoint ($t_{\text{phase}} = T$), the trajectory represented by the spline ends with zero velocity by design ($\dot{\mathbf{f}}(T) = 0$), which further ensures the autonomous system halts when reaching this last point. Therefore, the dot product term $\mathbf{\nabla d}(\mathbf{x})^\top \dot{\mathbf{f}}(T)$ is zero.
    \item At the \textbf{start point} ($t_{\text{phase}}=0$), we always have $\mathbf{\nabla d}(\mathbf{x})^\top \dot{\mathbf{f}}(0) \leq 0$. Otherwise,  it would mean the trajectory initially moves into the half-space containing $\mathbf{x}$ and the start point $\mathbf{f}(0)$ is not the closest point of $\mathbf{x}$ on the entire trajectory. 
\end{itemize}
Since the second term is non-positive in all cases, we confirm that \(\dot{V}(\mathbf{x}) \le 0\), ensuring global stability. Furthermore, \(\dot{V}(\mathbf{x}) = 0\) only when \(d(\mathbf{x}) = 0\), which implies that the system asymptotically converges to the reference trajectory.
}

To achieve a balance between guiding the system toward the desired trajectory and ensuring smooth motion along it, we introduce an inverse barrier function to modulate the influence of the gradient term with
\begin{equation}
    \beta = \frac{1}{1+\lambda \ d(\mathbf{x})}, \quad
    \alpha = 1 - \beta,
\end{equation}
where \( \lambda>0 \) is a parameter to balance the two terms. In essence, larger values of the distance field $d(\mathbf{x})$ correspond to regions farther from the trajectory, prompting the dynamical system to exert a stronger influence to steer the state back toward the trajectory. Conversely, smaller distance values indicate proximity to the trajectory, causing the system to reduce its corrective force and allow for smoother motion along the path. \textcolor{black}{The parameters, $\alpha$ and $\beta$ in balancing attraction and progression terms can also be integrated with the phase variable
$t_\text{phase}$ for more complex behaviors.}

\section{Experiments} 

\begin{table}[t]
\centering
\caption{Average reconstruction error for different basis functions.}
\label{tab:compare}
\Huge
\renewcommand{\arraystretch}{1.5} 
\resizebox{\linewidth}{!}{
\begin{tabular}{|c|c|c|c|c|c|}
\hline
\textbf{Method} & \textbf{3} & \textbf{7} & \textbf{12} & \textbf{17} & \textbf{22} \\ \hline
\textbf{Piecewise}  & $7.97\pm2.76$ & $3.40\pm1.10$ & $1.97\pm0.63$ & $1.39\pm0.44$ & $1.08\pm0.35$ \\ \hline
\textbf{B.P.}  & $6.18\pm3.65$ & $0.96\pm0.60$ & $0.28\pm0.13$ & $0.14\pm0.06$ & $0.10\pm0.04$ \\ \hline
\textbf{RBF}  & $16.51\pm4.35$ & $1.79\pm0.61$ & $0.26\pm0.08$ & $0.12\pm0.04$ & \textbf{$\mathbf{0.05\pm0.02}$} \\ \hline
\textbf{Fourier}  & \textbf{$\mathbf{5.18\pm3.52}$}  & \textbf{$\mathbf{0.56\pm0.33}$} & \textbf{$\mathbf{0.19\pm0.08}$} & \textbf{$\mathbf{0.09\pm0.04}$} & \textbf{$\mathbf{0.05\pm0.02}$} \\ \hline
\textbf{Q.S.} (ours)  & $6.18\pm3.65$  & $0.88\pm0.53$ & $0.23\pm0.10$ & $0.11\pm0.04$ & $0.06\pm0.02$ \\ \hline
\end{tabular}
}
\end{table}

We evaluated the proposed approach through numerical simulations and real-world robotic experiments. Our evaluation is guided by the following key questions:
\textbf{Q1}: How efficiently can quadratic splines capture and reproduce complex motion patterns?  
\textbf{Q2}: How do the analytical distance and gradient fields contribute to motion generation in autonomous systems?
\textbf{Q3}: How robust and generalizable is the approach in handling perturbations and enabling adaptive motion responses? 
We address each of these questions through detailed analysis and discussion of our experimental results.


\subsection{Numerical Experiments}
We first assess how well quadratic splines encode trajectories through numerical experiments on the LASA hand-written dataset.

\textbf{Metrics}. The quality of reconstruction is measured using the average $\ell_2$ norm error
\begin{equation}
    e = \frac{1}{N}\|\bm{\Phi}\bm{w} - \bm{P}\|,
\end{equation}
where \(\bm{\Phi}\bm{w}\) represents reconstructed points on the spline, \(\bm{p}\) is a vector containing the original points of the trajectory and $N$ is the number of points of \(\bm{P}\). 

\begin{figure}[t]
    \centering
    \includegraphics[width = 1.0\linewidth]{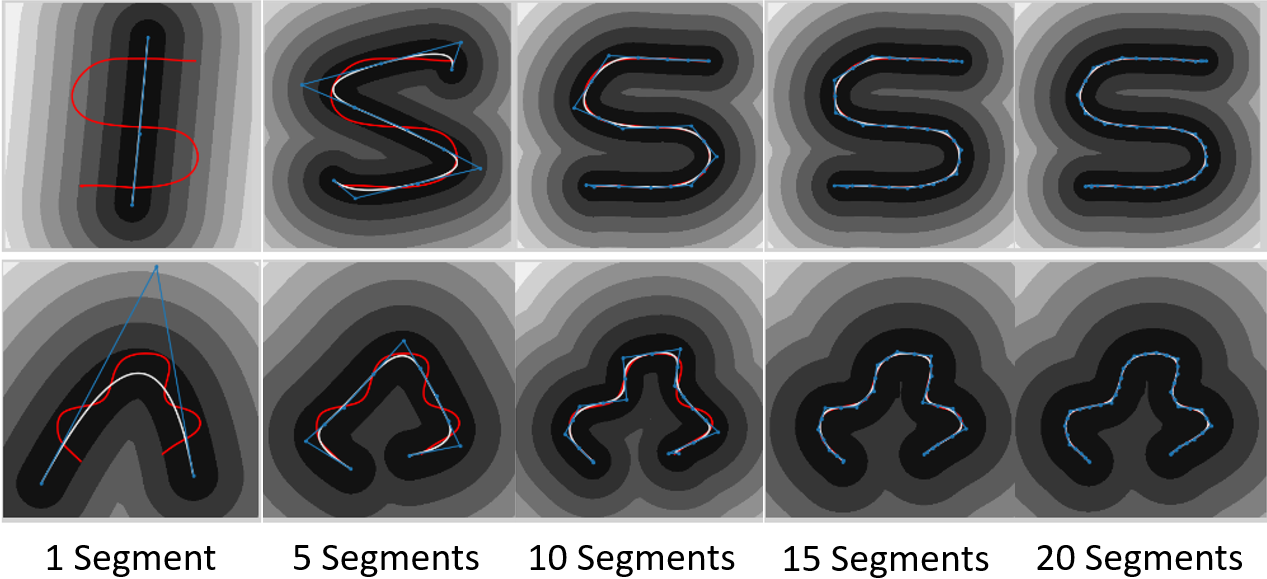}
    \captionof{figure}{The accuracy of quadratic splines in modeling trajectories improves as the number of segments increases.  
    The red curve represents raw data from the LASA dataset, while the white curve shows trajectories encoded using quadratic splines.  
    The blue components illustrate the superposition weights, and the colormap in the background represents the retrieved distance field.}
    \label{fig:Nsegments}
    \vspace{-3mm}
\end{figure}

\begin{figure}[t]
    \centering
    \includegraphics[width = 1.0\linewidth]{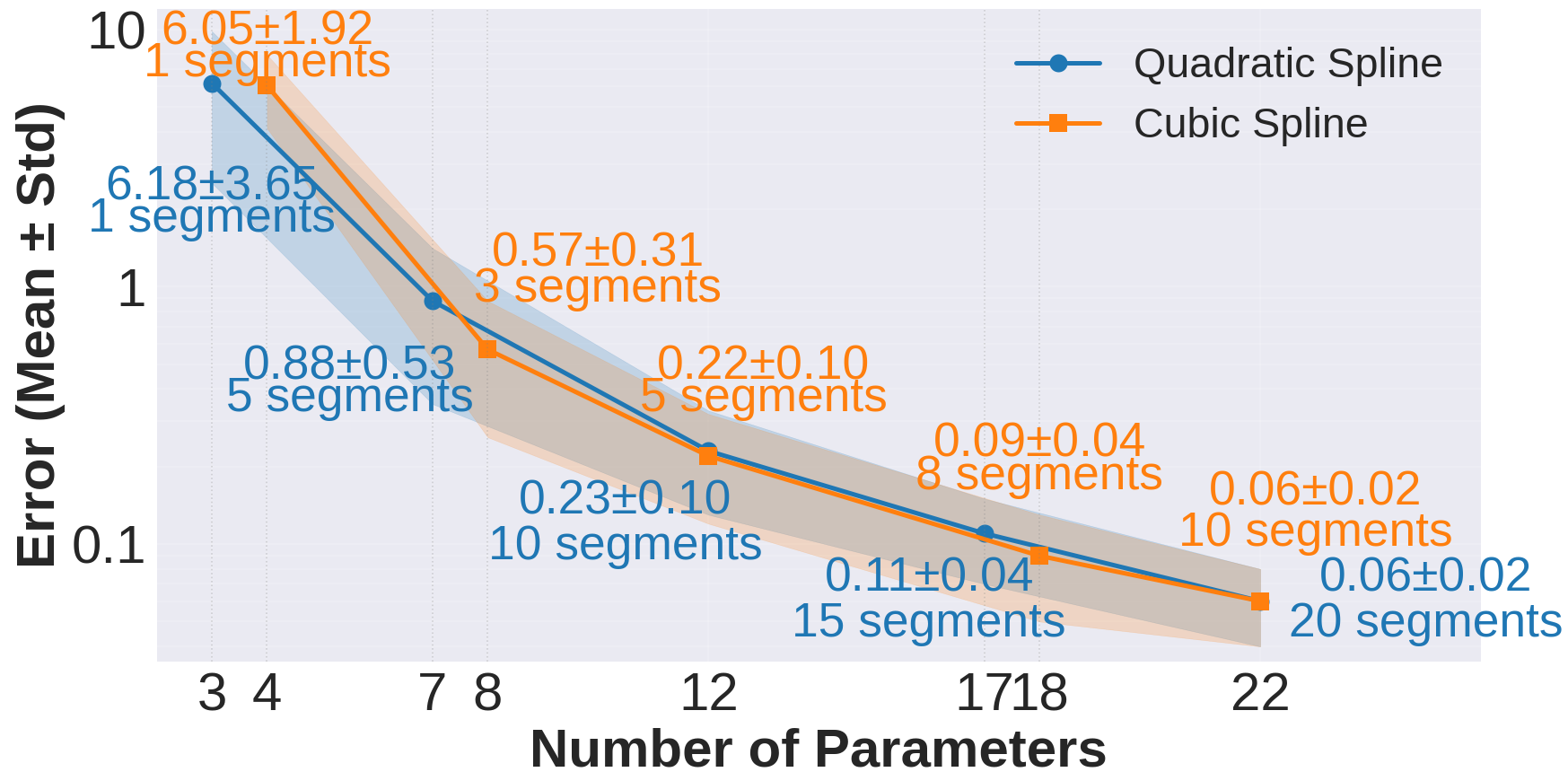}
    \captionof{figure}{\textcolor{black}{Comparison between quadratic and cubic splines in terms of trajectory encoding accuracy.}}
    \label{fig:quadratic_vs_cubic}
    \vspace{-4mm}
\end{figure}

\begin{figure}[t]
    \centering
    \includegraphics[width = 1.0\linewidth]{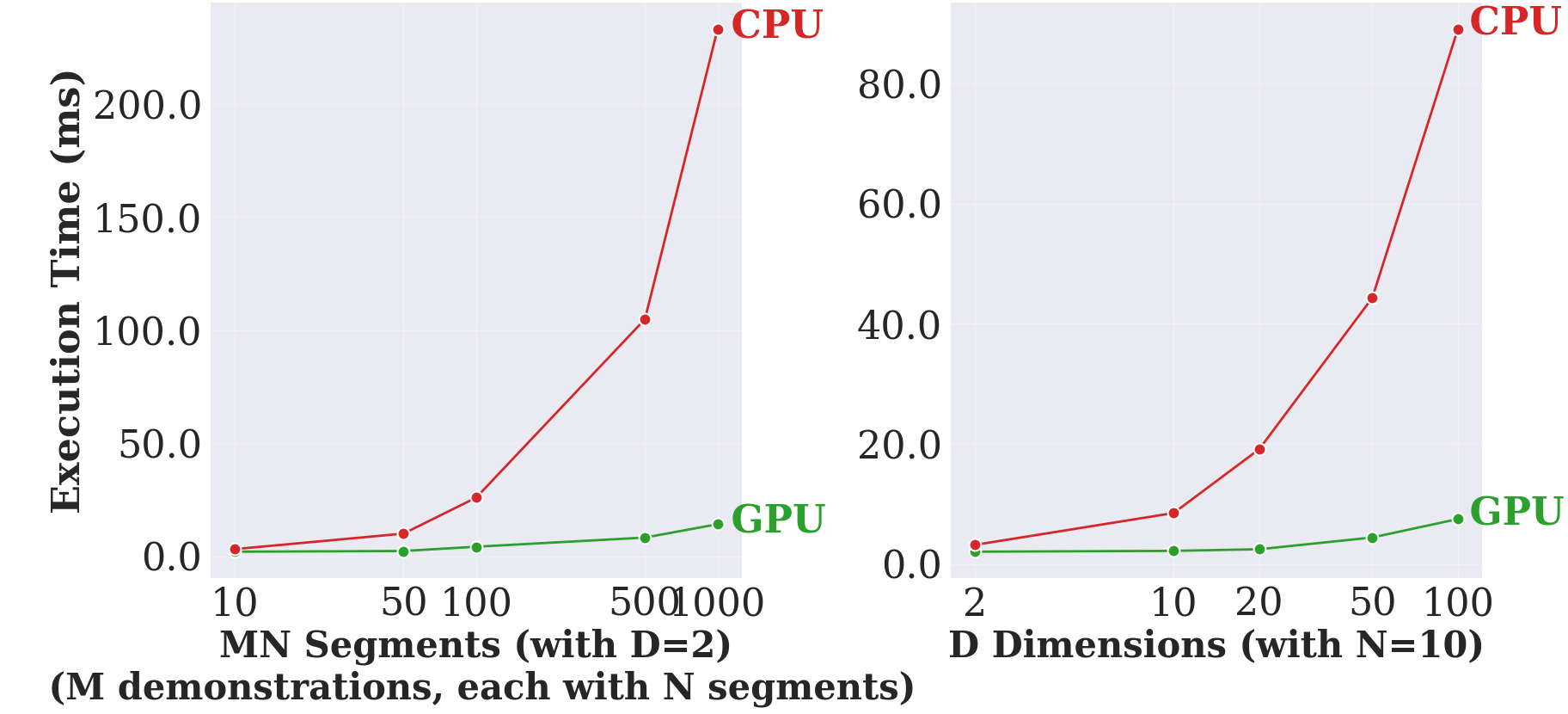}
    \captionof{figure}{\textcolor{black}{Computation time for various numbers of segments (or demonstrations) and dimensions.}}
    \label{fig:computation_time}
    \vspace{-6mm}
\end{figure}

\textbf{Accuracy}. To answer \textbf{Q1}, we compare the quadratic splines (Q.S.) with other basis functions, including piecewise constant basis functions, a single high-order Bernstein polynomial (B.P.), radial basis functions (RBF), and Fourier basis functions. For a fair comparison, we encode the trajectory using the same number of parameters. A piecewise quadratic spline with $N$ segments has $N+2$ parameters in total, so we use $N+2$ basis functions for each comparison method. Experimental results are reported in Table \ref{tab:compare}.

Although the quadratic spline only uses piecewise quadratic polynomials for trajectory encoding, it achieves comparable performance with other commonly used movement primitives. This is a benefit from the flexibility provided by the spline, which allows local adjustments to be made without affecting the entire trajectory (see Figure \ref{fig:Nsegments}). The piecewise line segments also enable distance field computation, however, it is limited by the low degree of the polynomials and the discontinuity of the derivatives. Therefore, the quadratic spline is preferable as it can encode trajectories expressively while providing an analytical form for distance computation. 
We also compare the results with cubic splines in Figure \ref{fig:quadratic_vs_cubic}, a concatenation of higher-order Bernstein polynomials which is also widely used in trajectory encoding. The cubic spline with $N$ segments has $2N+2$ parameters in total in the case of $C^1$ continuity. While cubic splines perform better with the same number of segments, they offer no significant advantage over quadratic splines for an equivalent number of parameters.

\textbf{Computation Time.}  
\textcolor{black}{We evaluate in Fig. \ref{fig:computation_time} the computational cost of our approach across varying numbers of segments and dimensions. The computational cost of fusing $M$ demonstrations, each with $N$ segments, is equivalent to processing a single trajectory composed of $MN$ segments.} We measure the time required to compute distances, gradients, and time phases for 2,500 points relative to a quadratic curve in parallel using PyTorch. Results are shown in Figure~\ref{fig:computation_time}. Notably, for typical robotic settings (e.g., $MN \leq 100$, $D \leq 10$), our method is highly efficient—requiring less than 5ms on a single NVIDIA GeForce RTX 3060 laptop GPU and less than 10\, ms on an AMD Ryzen 7 6800H CPU. Even with significantly larger $MN$ and $D$, the method remains efficient, particularly on the GPU. Further performance gains are possible due to the method’s simplicity and ease of implementation.

\textbf{Flexibility and Stability. }To answer \textbf{Q2}, we evaluate the dynamical system derived from the distance field by analyzing its behavior with varying \(\lambda\) values (Figure \ref{fig:ds}). A larger \(\lambda\) increases the influence of the distance field, enhancing resistance to disturbances. In contrast, smaller \(\lambda\) values allow for smoother curvature following, resulting in less rigid trajectory adherence. Adjusting \(\lambda\) directly impacts the system's responsiveness, stability, and convergence.
The results demonstrate the flexibility of the distance field representation in dynamically shaping autonomous system behaviors. Figure \ref{fig:ds} shows results on various shapes from the LASA dataset with \(\lambda = 3.0\).

\begin{figure}[t] 
    \centering \includegraphics[width=.9\linewidth]{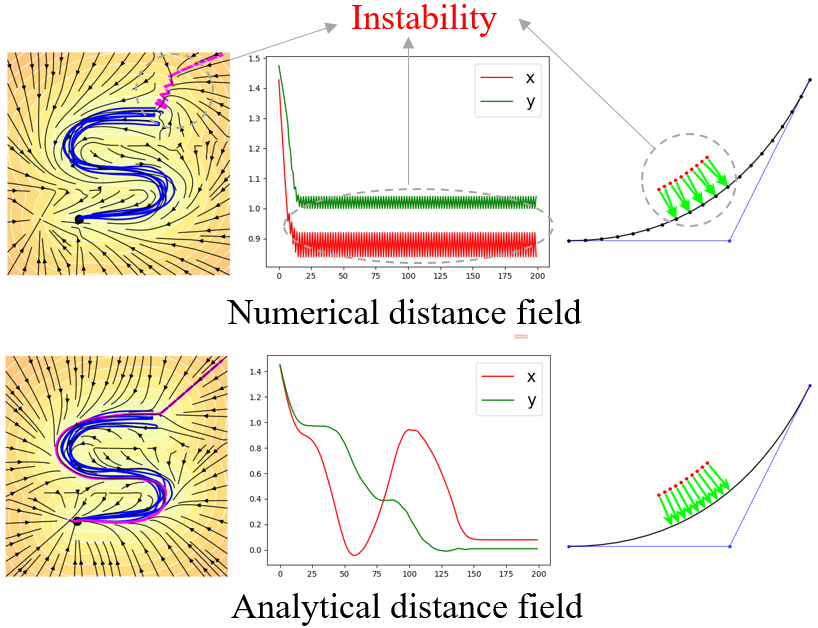} 
    \captionof{figure}{Dynamical systems using either numerical or analytical distance fields. The left plots show the produced trajectory. The right plots show the instability caused by discontinuities in the numerical gradient, leading to an unstable dynamical system.} 
    \label{fig:comparison_numerical_gradients}
\end{figure}

To demonstrate the importance of using an analytical distance gradient in the design of globally stable dynamical systems, we compare our method with an alternative that uses a numerically computed distance field. The numerical field is obtained by uniformly sampling points along the trajectory and finding the closest point for each query. The results are shown in Figure~\ref{fig:comparison_numerical_gradients}. The comparison reveals that numerical distance fields can introduce instability, as their gradients may not remain perpendicular to the trajectory, causing the dynamical system to exhibit jerky motion or become stuck. In contrast, our analytically derived gradients ensure smooth, stable behavior throughout the trajectory.


\begin{table}[t]
    \centering
    \begin{tabular}{ccc}
        \includegraphics[width=0.3\linewidth]{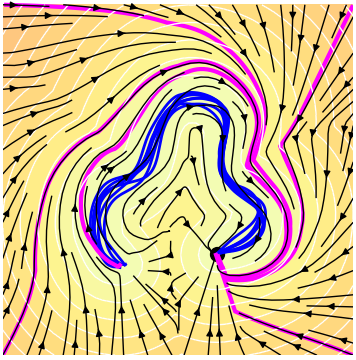} &
        \includegraphics[width=0.3\linewidth]{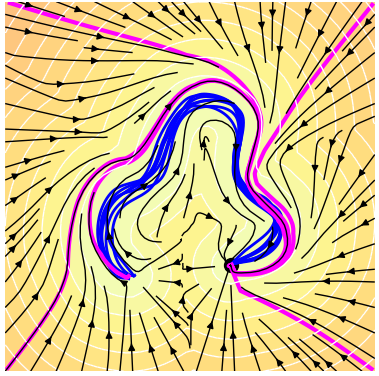} &
        \includegraphics[width=0.3\linewidth]{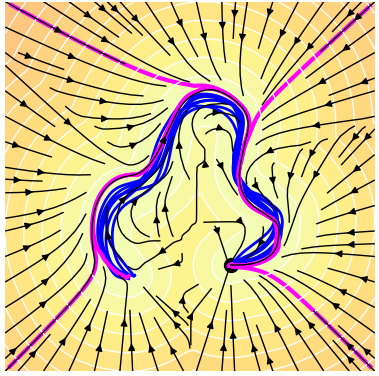} \\
        $\lambda = 0.5$ & $\lambda = 1.0$ & $\lambda = 3.0$ \\
        
        \includegraphics[width=0.3\linewidth]{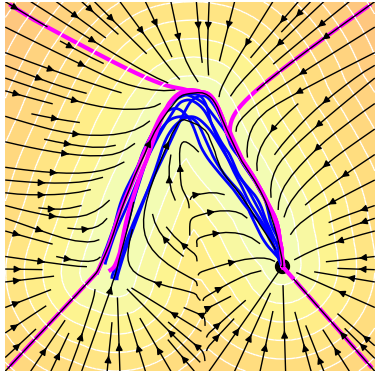} &
        \includegraphics[width=0.3\linewidth]{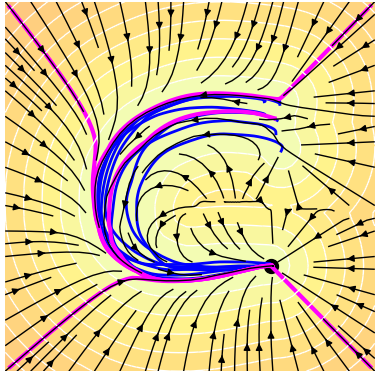} &
        \includegraphics[width=0.3\linewidth]{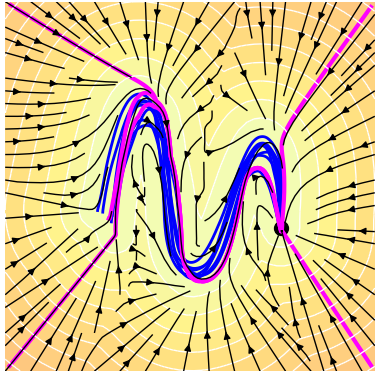} \\
        Angle & C-shape & Sine \\
        
        \includegraphics[width=0.3\linewidth]{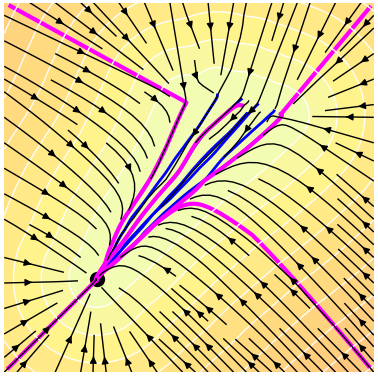} &
        \includegraphics[width=0.3\linewidth]{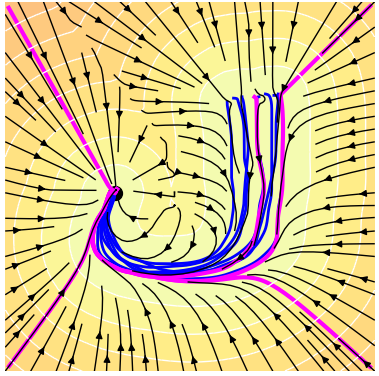} &
        \includegraphics[width=0.3\linewidth]{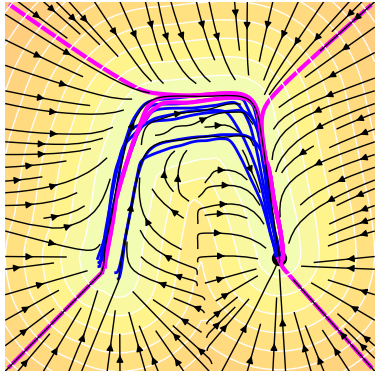} \\
        Line & J-shape & Trapezoid 
    \end{tabular}
    \captionof{figure}{Dynamical systems derived from the distance field. Blue curves represent demonstrations. Purple curves depict generated paths starting from the 4 corners of the image. 
    \emph{Top row:} Effect of varying $\lambda$. \emph{Other rows:} Results for different trajectories from the LASA dataset.}
    \label{fig:ds}
    \vspace{-6mm}
\end{table}

\begin{figure}[t]
    \centering
    \includegraphics[width = 1.0\linewidth]{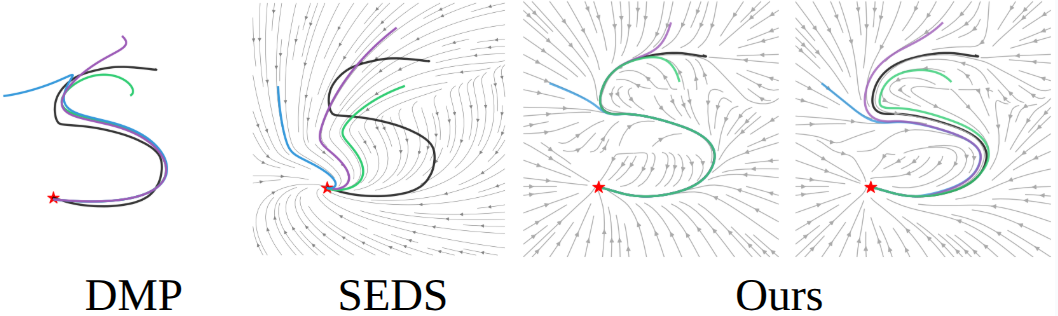}
    \captionof{figure}{\textcolor{black}{Qualitative comparison with baseline approaches. Black curves represent the original demonstration, while colored curves show the reproductions.}}
    \label{fig:comparison}
    \vspace{-3mm}
\end{figure}

\textcolor{black}{\textbf{Comparison with Baselines.} We compare our method against two widely-used approaches: Dynamic Movement Primitives (DMP) \cite{ijspeert2013dynamical} and Stable Estimator of Dynamical Systems (SEDS) \cite{khansari2011learning}. Our approach offers distinct advantages in terms of model capability and computational efficiency. Specifically,  DMP and SEDS rely on a learned model of the system's dynamics (e.g., a forcing term profile encoded with basis functions for DMP, or a Gaussian mixture model for SEDS). 
SEDS formulates the autonomous system as a constrained optimization problem. The constructed vector fields predominantly guide trajectories toward a single attractor, making it unsuitable for modeling and reproducing elaborate paths that would temporarily move away from the goal. 
It also becomes computationally challenging to apply this approach to high-dimensional spaces.
In contrast, our method constructs the dynamics directly from the geometry of the reference path (without learning a model) by superposing normal attraction and tangential progression vectors.
DMP is based on the advancement of a phase variable (reparameterization of time), instead of an autonomous system. It is also typically designed to learn from a single demonstration. In contrast, our approach constructs an autonomous system that naturally accommodates multiple demonstrations by fusing their respective distance fields through a simple and efficient minimum operation. 
Additionally, our method offers highly intuitive and modular control over the robot's behavior. A single parameter directly balances the influence of the attraction term (how strongly we need to come back to the path if a perturbation occurs) and the progression term (how to move along the path), which can be easily balanced (see the last two plots of Fig.~\ref{fig:comparison}). In contrast, the parameters in DMP and SEDS are often embedded within the learned models, making them less modular.
}
\subsection{Robot Experiments}
To answer \textbf{Q3}, we carry out three experiments with a 7-axis Franka robot:

\subsubsection{Disturbance Handling} 
\begin{figure}[t]
    \centering
    \includegraphics[width=0.8\linewidth]{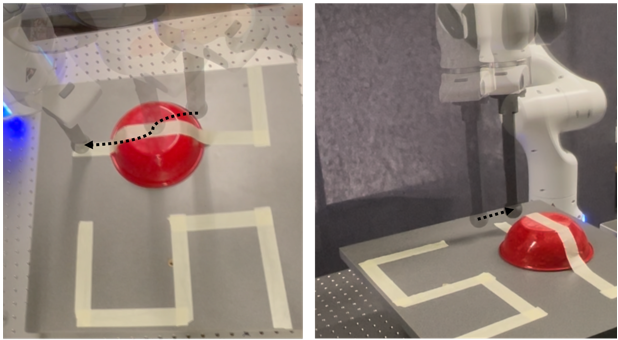}
    \includegraphics[width=0.8\linewidth]{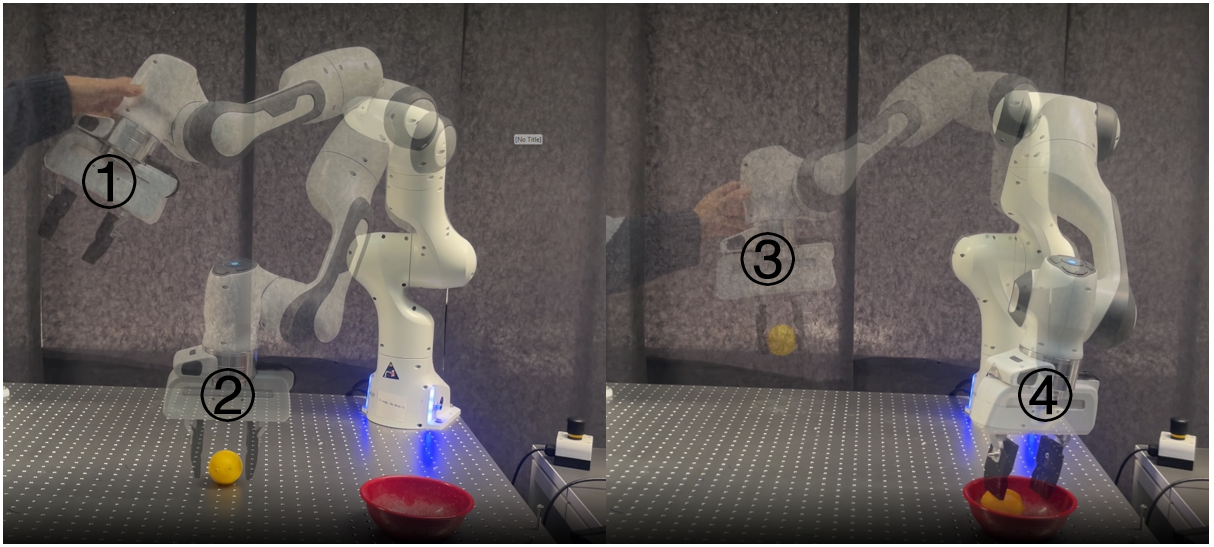}
    \includegraphics[width=0.8\linewidth]{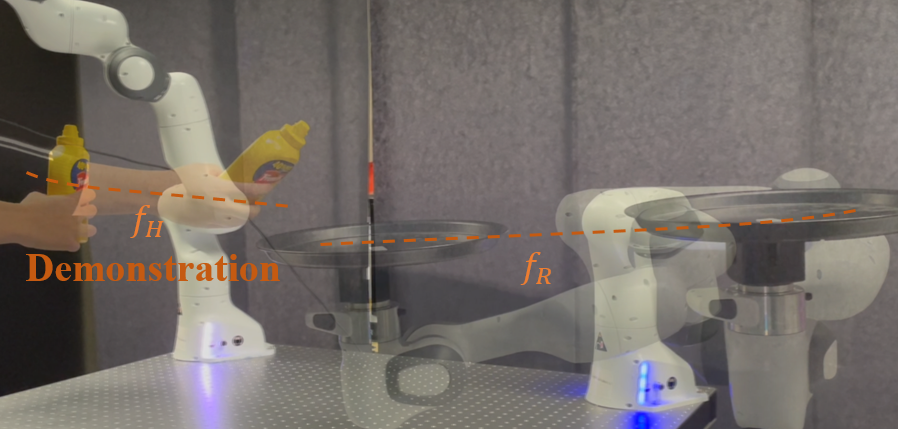}    

    \caption{\textcolor{black}{Real-world experiments. \textit{Top:} Disturbance handling. \textit{Middle:} Pick and place. \textit{Bottom:} Human-robot collaboration.}
    }
    \label{fig:exp_combined}
    \vspace{-3mm}
\end{figure}

In this experiment, the robot was tasked to maintain two demonstrated S-shaped and L-shaped trajectories in task space with the dimension $D=3$, as shown in Figure \ref{fig:exp_combined}-\emph{top}. Disturbances are introduced by a person physically moving the robot. In the left panel, disturbances were introduced along the trajectory, and the robot remains stationary after the disturbance. In the right panel, disturbances caused displacement along the normal direction of the reference trajectory; \textcolor{black}{In response, the manipulator returned to the closest point on the reference trajectory, demonstrating its adaptability to unexpected deviations, facilitated by the distance field.}

\textcolor{black}{
\subsubsection{Pick and Place}
In this task, the robot was required to pick up an object from the table and place it into a bowl (shown in Figure \ref{fig:exp_combined}-\emph{middle}). The demonstrated trajectory is defined in joint space with dimension $D\!=\!7$. During execution, we manually disturbed the robot by physically dragging it from the reference path. At each iteration of the control loop, the robot estimates its task phase and the closest point on the demonstrated trajectory, which defines the dynamical system to return to the desirable positions. The robot is controlled by a torque controller in a high-compliance mode, allowing us to apply external disturbances easily. IN the experiment, the gripper actions were triggered at predefined segments of the trajectory; however, this process could be integrated more dynamically into the overall framework by estimating the task phase online.
}

\subsubsection{Human-Robot Collaboration}  

\textcolor{black}{
We demonstrate our approach in a dynamic human-robot handover task. A single demonstration is first recorded, encoding both the human's hand trajectory in task space, \(f_H(t)\), and the robot's synchronized joint-space motion, \(f_R(t)\), as quadratic splines. During interaction, a camera tracks the human's current hand position, \(\mathbf{x}_H\). Our method is used here to project this position onto the learned human path \(f_H\) to continuously infer the task's current phase, \(t_{\text{phase}}\). This phase variable then directly drives the robot, which is commanded to the corresponding point on its own trajectory, \(\mathbf{f}_R(t_{\text{phase}})\). This coupling creates a fluid and anticipatory interaction where the robot dynamically synchronizes its timing with the human. It robustly adapts to unpredictable movements and sensor noise, ensuring a successful and efficient handover, as the system robustly handles unpredictable human timing and noisy sensor data. This adaptive behavior is illustrated in Figure~\ref{fig:exp_combined}-\textit{bottom} and in the supplementary video.
}

\section{Discussion and Conclusion}

We have demonstrated that trajectories encoded using quadratic splines can be systematically converted to implicit distance fields, which can further be converted to stable autonomous systems. The key approaches described in this paper involve the least-squares method to represent a trajectory as a quadratic spline, and solving a cubic equation for the corresponding distance field, which are very simple, straightforward and intuitive. The approach is adaptable to disturbances, dynamic environments and high-dimensional problems. While our primary focus has been on the adaptive and robust reproduction of demonstrations, the integration of movement primitives with distance fields opens new avenues for broader applications.

\textbf{Limitations}. In the approach presented in this paper, we re-estimate a phase/time variable t at each step without taking into account the previous estimates of t, in order to obtain a time-independent autonomous system. This design introduces sensitivity to initial conditions: small variations in the starting point may cause the system to converge to different segments of the trajectory. While acceptable in some scenarios, this behavior may be undesirable in applications requiring more deterministic behavior. Future work could investigate how the trace of intermediary phase/time variable estimates could be exploited to provide control commands as a trade-off between a fully myopic and a time-dependent system, similarly to model predictive control. Another possible limitation is that the encoded and retrieved path only has \(C^1\) continuity. While this is good enough in many applications, further extensions could be considered by doubling the dimensionality of the trajectories by representing both the position and velocity profiles with the quadratic spline, which would allow the problem to be extended to acceleration commands, resulting in paths with \(C^2\) continuity. This extension would also enable the handling of trajectories that cross in position space but that do not cross in the full state space.

\textbf{Future work}. One promising extension for future work is the formulation of optimal control problems to learn the spline keypoints, whose results are then re-interpreted as distance fields. 
Beyond deterministic representations, distance fields could also be extended into a probabilistic framework capturing uncertainty in trajectory execution. By modeling the distribution of distance values, this approach could be used to enhance motion planning under uncertainty, enabling safe and adaptive behaviors in unstructured environments. Furthermore, distance fields could be learned directly from multimodal inputs, such as images, videos, or vision language models, allowing robots to infer motion trajectories from high-level visual or textual descriptions. These extensions illustrate the versatility of distance fields as a motion representation module that can be integrated into many learning, planning, control, and optimization problems.





\section{Acknowledgments}
This work was supported by the China Scholarship Council (No. 202204910113), 
and by the State Secretariat for Education, Research and Innovation in Switzerland for participation in the European Commission's Horizon Europe Program through the INTELLIMAN project (\url{https://intelliman-project.eu/}, HORIZON-CL4-Digital-Emerging Grant 101070136) and the SESTOSENSO project (\url{http://sestosenso.eu/}, HORIZON-CL4-Digital-Emerging Grant 101070310). We thank Yan Zhang for assistance with the robot experiments and Martin Schonger for insightful discussions regarding dynamical systems.

\bibliographystyle{ieeetr}
\bibliography{references}

\end{document}